%% file: main.tex
\pdfoutput=1

\documentclass[11pt]{article}

\usepackage[final]{acl}

\usepackage{times}
\usepackage{latexsym}
\usepackage{amssymb}
\usepackage[T1]{fontenc}
\usepackage{makecell}
\usepackage[utf8]{inputenc}

\usepackage{amssymb}
\usepackage{xcolor}
\usepackage{mathtools}
\usepackage{amsmath}
\usepackage{amsfonts}

\usepackage{times}
\usepackage{latexsym}
\usepackage{graphicx}
\usepackage{multirow}
\usepackage{multicol}
\usepackage{algorithm}
\usepackage{algpseudocode}
\usepackage{booktabs}
\usepackage{xcolor}
\usepackage{microtype}
\definecolor{custompink}{rgb}{1, 0.6549, 0.8627}  
\definecolor{customblue}{rgb}{0.498, 1, 1}  

\usepackage{inconsolata}

\usepackage{graphicx}
\usepackage{multicol}
\usepackage{multirow}
\usepackage{booktabs}

\title{Exploring Iterative Controllable Summarization \\with Large Language Models}

\author{
    \makecell{Sangwon Ryu$^{1}$, Heejin Do$^{3}$, Daehui Kim$^{4}$, \\Hwanjo Yu$^{1,2}$, Dongwoo Kim$^{1,2}$, Yunsu Kim$^{5}$, Gary Geunbae Lee$^{1,2}$, Jungseul Ok$^{1,2}$} \\
  \centering
  \begin{tabular}[t]{c}
    $^{1}$GSAI, POSTECH \quad
    $^{2}$CSE, POSTECH \\
    $^{3}$ETH AI Center \quad
    $^{4}$AI future lab, KT \quad
    $^{5}$LILT\\
    \texttt{\small{\{ryusangwon, hwanjoyu, dongwoo.kim, gblee, jungseul\}@postech.ac.kr}}  \\
    \texttt{\small{heejin.do@ai.ethz.ch}\quad{daehui.kim@kt.com}\quad{yunsu.kim@lilt.com}}
  \end{tabular}
}

\begin{document}
\maketitle

\input{content/abstract}

\input{content/introduction}

\input{content/related}

\input{content/method}

\input{content/experiment}

\input{content/result}

\input{content/discussion}

\input{content/conclusion}

\input{content/limitation}

\input{content/ethics}

\input{content/acknowledge}

\bibliography{custom}

\input{content/appendix}

\end{document}

%% file: content/abstract.tex
\begin{abstract}

Large language models (LLMs) have demonstrated remarkable performance in abstractive summarization tasks. However, their ability to precisely control summary attributes (e.g., \textit{length} or \textit{topic}) remains underexplored, limiting their adaptability to specific user preferences. 
In this paper, we systematically explore the controllability of LLMs. To this end, we revisit summary attribute measurements and introduce iterative evaluation metrics, \textit{failure rate} and \textit{average iteration count}, to more precisely evaluate controllability beyond assessment of errors. Our findings show that LLMs struggle more with numerical attributes than with linguistic attributes. To address this challenge, we propose a guide-to-explain framework (GTE) for controllable summarization. GTE enables the model to identify misaligned attributes in the initial draft and guides it to self-explain errors in the previous output. By encouraging reflection on attribute misalignment, GTE generates well-adjusted summaries that satisfy the desired attributes with robust effectiveness while requiring surprisingly fewer iterations than other iterative approaches.

\end{abstract}

%% file: content/introduction.tex
\section{Introduction}

Large language models (LLMs) have demonstrated superior performance in text summarization, outperforming encoder-decoder models by generating more contextually appropriate and natural summaries \cite{goyal2023news, zhang2023benchmarking, pu2023summarization, ryu24_interspeech}.
However, given the diversity of individual preferences for summary styles, it is essential to generate summaries tailored to specific user needs \cite{zhang-etal-2023-macsum}. For example, some users may prefer topic-focused summaries or wish to retain exact phrases. Although LLMs excel at generating fluent summaries, their ability to precisely control attributes remains underexplored \cite{liu-etal-2024-benchmarking}, limiting their adaptability to diverse user preferences.
Typical requests can be ambiguous, such as \textquotedblleft summarize in 3 sentences\textquotedblright{} or \textquotedblleft generate a highly extractive summary\textquotedblright{}. Sentence lengths can vary significantly, and vague terms such as \textquotedblleft highly\textquotedblright{} hinder reliable instruction-following and complicate evaluating whether the instructions are properly satisfied (Figure~\ref{fig: ambiguous}).

\input{fig/fig-ambiguous}

Therefore, we systematically explore the controllability of LLMs. We begin by revisiting the measurements for four key attributes: \textit{extractiveness}, \textit{length}, \textit{topic}, and \textit{speaker}, and refine them for more precise measurement.
For length, instead of dividing summaries into fixed-length buckets and computing differences across them, which overlooks variations within each bucket, we directly measure the gap between the generated and target length. Moreover, rather than relying solely on word presence as in previous strategies for measuring \textit{topic}- or \textit{speaker}-focused summaries, we adopt embedding-based similarity to incorporate semantic information into the measurements.
With more precise attribute measurements in place, we next investigate how reliably LLMs can control these attributes. 
To fully explore LLM controllability, we evaluate whether LLMs can accurately control specified attributes through iterative refinement. Even if initial attempts fail, we test whether they can eventually succeed without external guidance.
To this end, we introduce two evaluation metrics: the \textit{failure rate}\textemdash{}the proportion of control failures within the maximum iterations\textemdash{}and the \textit{average iteration count} until successful control. In Section~\ref{sec: LLM controllability}, we show that while LLMs excel at controlling linguistic attributes such as \textit{topic} and \textit{speaker}, they struggle significantly with numerical attributes such as \textit{extractiveness} and \textit{length}. We assume that, unlike linguistic attributes, which rely on semantic coherence, numerical ones require adherence to strict structural constraints, making fine-grained control challenging.

To address this challenge, we introduce a guide-to-explain (GTE) framework, a self-correction-based approach that enables precise attribute control through LLMs without relying on additional attribute-specific training.
Although self-correction approaches have been employed to evaluate and revise their own outputs for complex tasks, we observe that LLMs often fail to reliably measure their output summary attributes. 
Therefore, we provides the model with guidance and encourages it to reflect on why its previous attempt was incorrect before regenerating the summary.
We first design a step-by-step attribute identification phase that instructs the model to identify misaligned attributes in its previously generated summary and then guides it to explain the rationale behind its errors.
Through self-reflection, the model corrects its prior mistakes and generates a well-aligned summary in the regeneration phase. By integrating a self-refinement strategy into controllable summarization, we improve the controllability of LLMs while enhancing summary quality.

We evaluate GTE on mixed-attribute summarization datasets, MACSum$_{Doc}$ and MACSum$_{Dial}$ \cite{zhang-etal-2023-macsum}.
Whereas standard iterative methods often prove ineffective for controllable summarization, our modified approach achieves reliable attribute control with minimal iterations and consistently adjusts attributes across data samples.
Furthermore, we demonstrate the high quality of the controlled summaries across multiple generic summarization evaluation metrics, including UniEval \cite{zhong-etal-2022-towards} and QuestEval \cite{scialom-etal-2021-questeval}.
Finally, we analyze whether LLMs can control multiple attributes simultaneously, revealing their difficulty in jointly managing correlated numerical attributes. Our contributions are as follows:

\begin{itemize}
    \item We systematically explore the controllability of LLMs in text summarization.
    \item We refine the measurement of summarization attributes and introduce \textit{iterative evaluation} metrics to evaluate LLM controllability.
    \item We propose a guide-to-explain (GTE) framework, which guides the model to explain its misalignments and effectively adjust attributes within just a few iterations.
\end{itemize}

%% file: fig/fig-ambiguous.tex
\begin{figure}[t]
\centering
\includegraphics[width=\linewidth]{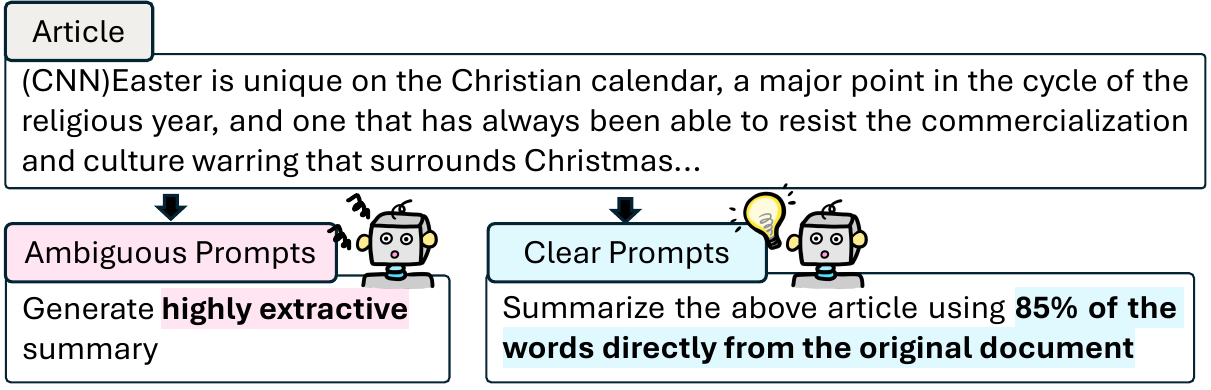}
\caption{Ambiguous instructions hinder LLMs' ability to follow control signals and complicate the evaluation process (e.g., how should \textquotedblleft highly\textquotedblright{} be judged in a generated summary?).}
\label{fig: ambiguous}
\end{figure}

%% file: content/related.tex
\section{Related work}

\paragraph{Controllable summarization} Controllable summarization has recently gained attention due to its practical applications \cite{zhong-etal-2021-qmsum, ahuja-etal-2022-aspectnews, maddela-etal-2022-entsum, mehra-etal-2023-entsumv2, xu-etal-2023-lmgqs, zhang-etal-2023-macsum, ribeiro-etal-2023-generating, retkowski2025}. Previous research has employed encoder-decoder models to control attributes \cite{fan-etal-2018-controllable, liu-chen-2021-controllable, dou-etal-2021-gsum, he-etal-2022-ctrlsum, mao-etal-2022-dyle, zhang-etal-2022-summn, goyal-etal-2022-hydrasum, vig-etal-2022-exploring, bahrainian-etal-2022-newts, liu-etal-2022-length, pagnoni-etal-2023-socratic, wang-etal-2023-instructive, urlana-etal-2024-controllable}. For example, CTRLSum \cite{he-etal-2022-ctrlsum} trains models by prepending a keyword sequence to the source document. Similarly, MACSum \cite{zhang-etal-2023-macsum} adopts prompt learning by prepending each attribute's value to the input using a combination of hard prompts and soft prefixes. HYDRASUM \cite{goyal-etal-2022-hydrasum} leverages a single encoder, multiple decoder framework with a mixture-of-experts approach, where the decoders share probabilities to effectively control the attributes.

Most controllable summarization research has relied on encoder-decoder frameworks. In addition, these methods often require attribute-specific training or custom datasets to control each attribute, limiting the flexibility of attribute manipulation. 
Therefore, we propose a generalizable approach that enables flexible attribute control without the need for tailored training, leveraging LLMs for controllable summarization \cite{tang-etal-2023-context, yuan2024followinglengthconstraintsinstructions, liu-etal-2024-benchmarking}.

\paragraph{Self-correction} Recently, self-correction approaches have been used to refine initial attempts at solving complex problems \cite{weng-etal-2023-large, shinn2023reflexionlanguageagentsverbal, self-refine-2023, dhuliawala-etal-2024-chain, gou2024critic}, mirroring human behavior. In summarization tasks, self-correction has been employed to enhance the overall quality of summaries \cite{zhang-etal-2023-summit, sun-etal-2024-prompt}. \citet{zhang-etal-2023-summit} utilizes iterative feedback from an evaluator to instruct ChatGPT to produce higher-quality summaries. Unlike prior work, we focus on generating summaries tailored to user preferences, which involve multiple factors to consider.

%% file: content/method.tex
\section{Attribute Measurement and Evaluation Framework for LLM Controllability}

We first analyze how each summarization attribute has traditionally been measured and redefine those that were not clearly defined. In particular, we refine linguistic attributes\textemdash{}often measured by word count, using embedding-based similarity. These refined measurements allow us to more accurately capture the attributes of generated summaries. Building on this, we propose iterative evaluation metrics to assess the controllability of LLMs\textemdash{}that is, their ability to precisely adjust attributes through multiple rounds of control.

\subsection{Revisiting attribute measurements for controllable summarization}\label{sec: attribute_measurement}

We revisit attribute measurement to quantify key attributes for controllable summarization: \textit{extractiveness}, \textit{length}, \textit{topic}, and \textit{speaker}. We observe that the measurements for certain attributes have not yet been clearly defined, leading to unclear criteria for evaluating controllability. Therefore, we outline our newly defined approach for attribute measurements below. 

\paragraph{Extractiveness} quantifies the degree of lexical overlap between a summary and its source document. a highly extractive summary is preferred when users need to retain the original context, such as in legal documents, whereas paraphrasing is often favored in general cases. Following the definition of \textit{extractiveness}, we measure the attribute as the proportion of words in the summary directly reused from the source text.

\paragraph{Length} refers to the number of words or sentences in the summary or the ratio of the summary's length to that of the original text. By controlling the length, the amount of information in the summary can be adjusted according to user preferences.
Prompts used in earlier work often specify a fixed number of sentences (e.g., "3 sentences"), but this approach fails to account for variations in sentence length and does not accurately reflect the summary's actual length \cite{goyal2023news, liu-etal-2024-benchmarking, yuan2024followinglengthconstraintsinstructions}. 
Since summary length may vary with document complexity~\cite{ryu-etal-2024-multi} and the original dataset~\cite{zhang-etal-2023-macsum} annotates length based on ratios, we use the length ratio rather than absolute length in our experiments. Further, prior studies~\cite{he-etal-2022-ctrlsum, zhang-etal-2023-macsum} evaluated length controllability by grouping outputs into discrete buckets and measuring differences across buckets. However, this approach overlooks variations within each bucket. Thus, we directly compare the generated summary length with the requested target length.

\paragraph{Topic} refers to generating a summary centered around one or more themes. Query-focused summarization (QFS), which generates summaries based on a specific query, and entity-based summarization, which focuses on a particular individual, are variations of topic-focused summarization. Most prior work has measured topic word frequency in summaries~\citet{fan-etal-2018-controllable, he-etal-2022-ctrlsum, zhang-etal-2023-macsum}. Similarly, most QFS methods have relied solely on ROUGE scores, evaluating generated summaries by comparing them to human-annotated references \cite{zhong-etal-2021-qmsum}.
However, even when topic words do not explicitly appear, a summary can still reflect the core context of the topic\textemdash{}especially in LLM-generated summaries, which tend to paraphrase content. 
Therefore, rather than simply counting word occurrences, we evaluate the semantic similarity between the summary and each topic-related word.
We compute the embedding similarity $\mathcal{B}$ between the topic word and each word in the summary $s$ as follows: $\frac{1}{n} \sum_{i \in s}\mathcal{B}(topic, word_{i})$, where $n$ is the number of words in the summary. If multiple topics $k$ are present, we use the average embedding similarity across all topics: $\frac{1}{k}\sum_{j \in k}\frac{1}{n} \sum_{i \in s}\mathcal{B}(topic_{j}, word_{i})$.

\paragraph{Speaker} refers to generating a summary that focuses on the utterances of a specific speaker within a long document or dialogue.
\citet{zhang-etal-2023-macsum} calculate the frequency of the speaker's spoken words appearing in the summary. Similar to \textit{topic} measurement, simply counting the proportion of words from a specific speaker's dialogue included in the summary does not fully capture semantic alignment. Therefore, we extract the speaker's utterances to construct a speaker set $\mathcal{U}$ and leverage BERTScore F1 \cite{zhang2020bertscore} to compute the embedding similarity between the summary $s$ and $\mathcal{U}$: BERTScore$(s, \mathcal{U})$.

\subsection{Iterative controllability evaluation}\label{sec: iterative_evaluation}

Building on these refined measurements of summary attributes, we introduce iterative evaluation metrics to assess whether LLMs can iteratively refine and adjust attributes over multiple revisions. Specifically, we introduce two metrics: (1) the \textit{failure rate}, proportion of cases in which the model reaches the predefined maximum number of iterations without achieving the desired modifications, and (2) the \textit{average iteration count} required for successful attribute control. We set the maximum number of iterations to 20 due to cost constraints.

\paragraph{Iteration threshold} We set attribute-specific thresholds and iteratively regenerate summaries until those thresholds are met. Each attribute is measured using the criteria outlined in Section \ref{sec: attribute_measurement} to determine its respective threshold. For \textit{extractiveness} and \textit{length}, we consider control successful if the attribute values fall within $\pm$5 of the target value. For \textit{topic} and \textit{speaker}, we use the minimum embedding similarity values of the reference summaries in the training dataset as thresholds to determine whether a summary is \textit{topic}-focused or \textit{speaker}-focused. Theses thresholds can be adjusted based on the strictness of the evaluation criteria. The distribution of the datasets used in our experiments is provided in Appendix~\ref{appendix:attribute_distribution}.

\input{fig/tab-llm_naive}

\paragraph{Label reinterpretation}
We use the two publicly available MACSum datasets \cite{zhang-etal-2023-macsum} for controllable summarization. 
However, existing labels are ambiguous, as the criteria are not numerically defined (e.g., how short must a summary be to qualify as short?).
We believe that such ambiguity may confuse LLMs, so we assign clear numerical values to each label. To provide detailed criteria, we reinterpret the labels based on the attribute distributions in each training set (see Appendix~\ref{appendix:attribute_distribution}). 
For \textit{extractiveness}, we define the labels as normal: 85\%, high: 90\%, and fully: 100\%, based on the average values.
For the \textit{length} attribute, we follow the annotation criteria of the MACSum dataset\textemdash{}short: 5–10\%, normal: 15–25\%, and long: 30–35\%\textemdash{}and set our target values to short: 7.5\%, normal: 15\%, and long: 32.5\%.
Importantly, our method generates summaries based on the specified numerical values, regardless of predefined labels.

\section{Analysis on Controllability of LLMs}\label{sec: LLM controllability}

\subsection{Iterative Evaluation on LLMs}

\input{fig/fig-self-reflection}
\input{fig/fig-main}

As research on leveraging LLMs for controllable summarization remains limited, we evaluate the controllability of various LLMs using the iterative evaluation method described in Section \ref{sec: iterative_evaluation}. We first provide an initial control prompt and generate a summary. If the generated summary fails to meet the specified attribute threshold, the result is fed back into the LLM’s input, prompting it to regenerate until the attribute is correctly controlled.
As shown in Table \ref{tab: llm_naive}, smaller-scale LLMs such as Phi-3-{\small{medium}} \cite{abdin2024phi} and Llama3-8B \cite{dubey2024llama}, partially control \textit{topic}, but fail to control \textit{extractiveness} and \textit{length}.
Similarly, large-scale LLMs such as Llama3-70B, GPT-3.5 \cite{brown2020language}, and GPT-4o \cite{achiam2023gpt} effectively control \textit{topic}, demonstrating low failure rates. However, they struggle with \textit{extractiveness} and \textit{length}, with failure rates of around 50\%. 
Notably, when initial attempts fail, even GPT-4o is unable to adjust after multiple iterations, ultimately reaching the maximum iteration limit, resulting in an iteration count of zero.
These findings suggest that generating summaries while controlling attributes remains challenging for LLMs, even with iterative attempts, especially for numerical attributes such as \textit{extractiveness} and \textit{length}.

\subsection{Self-correction for controllable summarization}

We evaluate whether LLMs can adjust summary attributes through self-correction, which has previously improved performance in generic summarization \cite{zhang-etal-2023-summit, sun-etal-2024-prompt}. However, unlike in generic summarization tasks, LLMs struggle to measure attributes, as shown in Figure~\ref{fig: self-reflection}. Specifically, they fail to accurately count words in either the source or the summary, making it infeasible for them to revise summaries to match target attribute values on their own.

\section{Guide-to-Explain (GTE)}

Therefore, we introduce a guide-to-explain (GTE) framework to control attributes via LLMs. As shown in Figure~\ref{fig: main}, the GTE consists of two key phases: step-by-step attribute identification and self-explanation guidance. Since LLMs struggle to reliably measure summary attributes on their own (Figure~\ref{fig: self-reflection}), we explicitly provide the attribute values and teach the model step by step how each attribute should be identified. We then guide the LLM to reflect by explaining the rationale behind its mistakes, enabling it to make appropriate adjustments in subsequent iterations.

\input{fig/tab-main_doc}

\input{fig/tab-main_dial}

\subsection{Step-by-step attribute identification} We first instruct the LLM to generate an initial summary $s'$ that reflects the specified attribute. If the LLM fails to control the attributes accurately, we provide step-by-step attribute identification (\texttt{SAI}) to guide the model on how to adjust them with tools. Since LLMs struggle with measuring numerical attributes such as \textit{extractiveness} or the \textit{length} ratio, we explicitly instruct the model on how to measure each attribute step by step, enabling it to revise its previously generated summary more precisely.

\subsection{Self-explanation guidance} After the identification phase, we provide self-explanation guidance (\texttt{SEG}) to the model, guiding the model to explain why it initially failed to control the attributes. This mirrors how humans solve complex problems by reviewing their mistakes to improve future responses. 
Building on this, in the next iteration, the document ($d$), initial instruction ($i$), and previously generated summary ($s'$) are provided as inputs, along with \texttt{SAI} and \texttt{SEG}. 
Although LLMs are known to struggle with number-related tasks \cite{akhtar-etal-2023-exploring, imani-etal-2023-mathprompter}, our guidance helps the model effectively control numerical attributes by self-explaining its miscalculations before generating summaries, especially when combined with the step-by-step attribute identification phase. We introduce GTE as a framework that integrates step-by-step attribute identification and self-explanation guidance.

\subsection{Overall process}
Figure~\ref{fig: main} illustrates in detail how the GTE framework operates. By receiving [$d$; $i$, $s'$; \texttt{SAI}; \texttt{SEG}] as input, the model first reflects on the reasons for its initial error before generating a revised summary.
If the revised summary still fails to satisfy the attributes, GTE repeats the process until the model generates an attribute-aligned summary. See Appendix~\ref{appendix:gte_prompts} for the detailed prompts.

%% file: fig/tab-llm_naive.tex
\begin{table}[t]
\centering
\scalebox{0.7}{
\begin{tabular}{l|c|c|c}
\toprule
\textbf{}          &  \textbf{Extractiveness} & \textbf{Length} & \textbf{Topic} \\ 
\midrule
Phi-3-{\small{medium}}       & 100.00\% /     $\circlearrowright$         & 100.00\% / $\circlearrowright$    & 38.08\% / 0.22     \\ 
Llama3-8B          & 100.00\% / $\circlearrowright$            & 100.00\% / $\circlearrowright$    & 57.14\% /  0.12    \\ 
Llama3-70B         & 49.91\% / 8.05             & 49.36\% /  8.24    & 0.00\% / 0.24   \\ 
GPT-3.5             & 49.73\% /  9.80            & 76.42\% / 0.00     & 0.00\% / 0.00  \\ 
GPT-4o              & 39.31\% /  6.63            & 69.84\% / 0.00     & 0.38\% / 0.02  \\
\midrule
\end{tabular}}
\caption{We evaluate the controllability of LLMs by iteratively testing their ability to accurately adjust specified attributes. The left number represents the averaged control \textit{failure rate}, and the right side denotes the \textit{average iteration count} for successful control.}\label{tab: llm_naive} 
\end{table}

%% file: fig/fig-self-reflection.tex
\begin{figure}[t]
\centering
\includegraphics[width=0.95\linewidth]{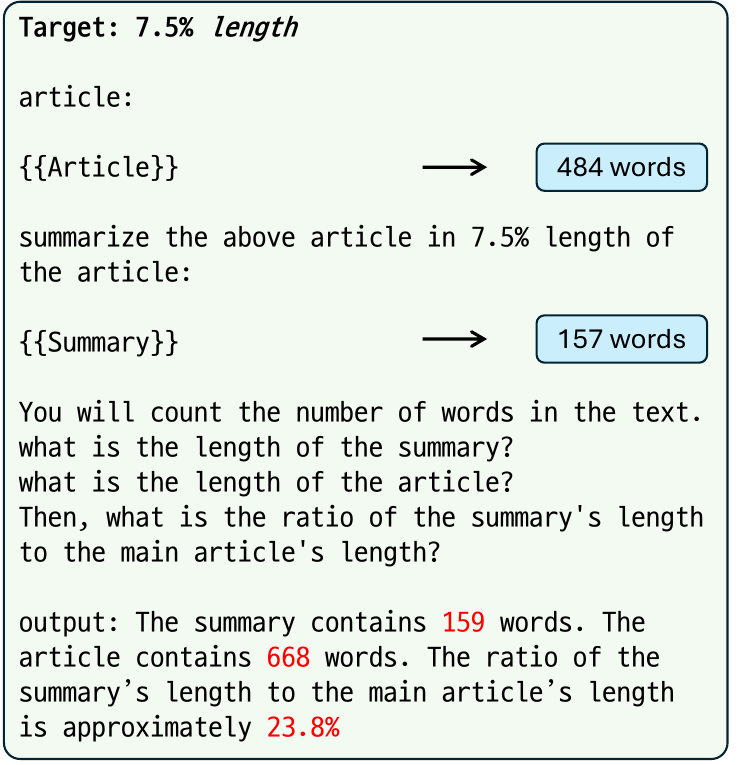}
\caption{LLMs show notable errors in word count estimation: for an article with 484 words and a summary with 157 words, the model predicts 668 and 159 words, respectively—revealing limitations in self-critique within controllable summarization.}
\label{fig: self-reflection}
\end{figure}

%% file: fig/fig-main.tex
\begin{figure*}
\centering
\includegraphics[width=\linewidth]{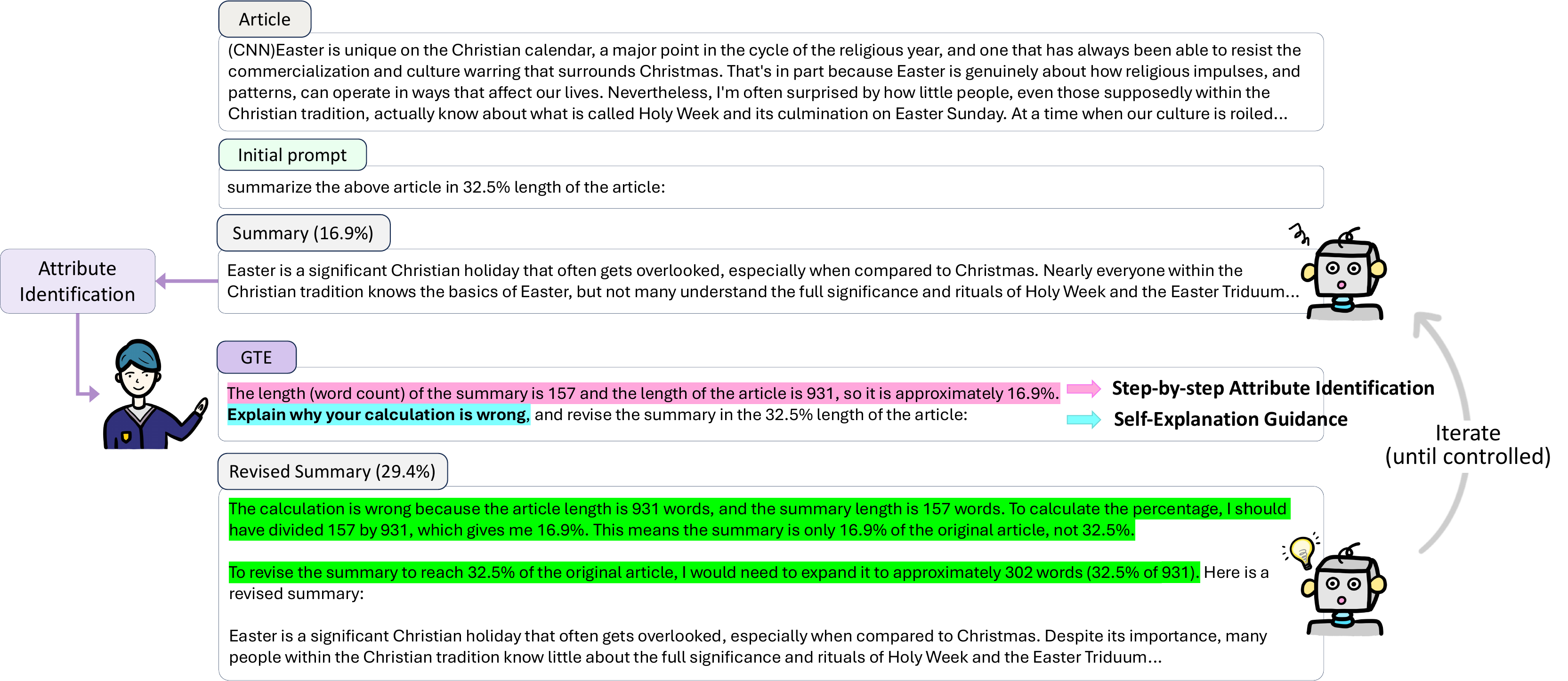}
\caption{Overview of guide-to-explain system (GTE). The pink parts ({$\color{custompink}\blacksquare$}) represent the step-by-step attribute-identification, and the blue parts ({$\color{customblue}\blacksquare$}) correspond to the self-explanation guidance.}
\label{fig: main}
\end{figure*}

%% file: fig/tab-main_doc.tex
\begin{table*}[]
\centering
\resizebox{1\textwidth}{!}{%
\begin{tabular}{l|cccc|cccc|c}
\toprule
\multirow{2}{*}{Model} & \multicolumn{4}{c|}{Extractiveness ($\downarrow$ / $\downarrow$)} & \multicolumn{4}{c|}{Length ($\downarrow$ / $\downarrow$)} & \multirow{2}{*}{Topic($\downarrow$ / $\downarrow$)}  \\ \cmidrule(lr){2-5} \cmidrule(lr){6-9} 
 & normal & high  & fully  & avg  & short  & normal  & long  & avg  &   \\ 
\midrule
Phi-3-{\small{medium}}-\texttt{Iter} & 100.00\% / $\circlearrowright$ & 100.00\% / $\circlearrowright$  & 100.00\% / $\circlearrowright$ & 100.00\% / $\circlearrowright$ & 100.00\% / $\circlearrowright$ & 100.00\% / $\circlearrowright$ & 100.00\% / $\circlearrowright$ & 100.00\% / $\circlearrowright$  & 38.08\% / 0.22  \\ 
Phi-3-{\small{medium}}-GTE & 100.00\% / $\circlearrowright$ &  100.00\% / $\circlearrowright$ &  100.00\% / $\circlearrowright$ &  100.00\% / $\circlearrowright$ & 100.00\% / $\circlearrowright$ &  100.00\% / $\circlearrowright$ & 100.00\% / $\circlearrowright$ &  100.00\% / $\circlearrowright$ & 37.97\% / 0.04  \\ 
\midrule
Llama3-8B-\texttt{Iter} & 100.00\% /  $\circlearrowright$  & 100.00\% / $\circlearrowright$ &  100.00\% / $\circlearrowright$ &  100.00\% / $\circlearrowright$ &  100.00\% / $\circlearrowright$  & 100.00\% / $\circlearrowright$ &  100.00\% / $\circlearrowright$ &  100.00\% / $\circlearrowright$ & 57.14\% / 0.12  \\ 
Llama3-8B-GTE  & 12.63\% / 3.52  &   11.63\% / 2.53 &   0.00\% / 1.46 &  11.70\% / 3.26 &   26.40\% / 3.08  &  10.92\% / 2.26 & 13.18\% / 3.85 &   14.99\% / 2.80 &  25.56\% / 0.91 \\ 
Llama3-70B-\texttt{Iter} &  54.82\% / 8.44 & 37.21\% / 7.47 &  2.70\% / 3.78 &  49.91\% / 8.05 & 18.40\% / 6.58 &  54.61\% / 10.42 &  67.44\% / 12.00 & 49.36\% / 8.24 & 0.00\% / 0.24   \\ 
Llama3-70B-\texttt{SAI} &  26.55\% / 6.57 &  18.60\% / 7.81 &  0.00\% / 1.86 &  24.14\% / 6.52 &  4.80\% / 5.42 &  2.73\% / 3.81 &   10.85\% / 4.84 &  5.12\% / 4.39 & 0.00\% / 0.10  \\ 
Llama3-70B-GTE  &  \textbf{0.21\%} / 3.28 &  \textbf{0.00\%} / 2.83 &  \textbf{0.00\%} / 1.50 &  \textbf{0.18\%} / 3.22 &  \textbf{0.00\%} / 1.10 &  \textbf{0.00\%} / 1.61 &  \textbf{2.32\%} / 3.14 &  \textbf{0.55\%} / 1.90 &  0.00\% / 0.01   \\ 
\midrule
GPT-3.5-\texttt{Iter} &  45.18\% / 9.80 &  60.47\% / 0.00 &  94.59\% / 0.00 &  49.73\% / 9.80 &  53.60\% / 0.00 &  80.89\% / 0.00 & 88.37\% / 0.00 &  76.42\% / 0.00 &  0.00\% / 0.00   \\ 
GPT-3.5-GTE &  17.56\% / 3.86 &   51.16\% / 5.00 &  67.57\% / 4.00 &  23.58\% / 3.90 &  5.60\% / 4.63 & 44.03\% / 6.62 &  78.29\% / 7.00 &  43.33\% / 5.95 &  0.00\% / 0.00  \\ 
GPT-4o-\texttt{Iter} &  34.69\% / 6.77 &  55.81\% / 0.00 &  78.38\% / 3.00 &  39.31\% / 6.63 &  72.00\% / 0.00 &  64.85\% / 0.00 &  79.07\% / 0.00 &  69.84\% / 0.00 &  0.38\% / 0.02   \\ 
GPT-4o-\texttt{SAI} &  35.12\% / 5.50 &  48.84\% / 15.50 &  62.16\% / 6.00 &  38.03\% / 6.13 &   60.00\% / 8.79 &  61.09\% / 9.40 &   78.29\% / 2.00 &  64.90\% / 8.60 &  0.00\% / 0.04   \\
GPT-4o-GTE  &  \textbf{0.00\%} / 2.76 &  \textbf{0.00\%} / 4.70 &  \textbf{0.00\%} / 2.03 &  \textbf{0.00\%} / 2.87 &  \textbf{0.00\%} / 1.20 &  \textbf{0.00\%} / 1.21 &  \textbf{0.00\%} / 1.96 &  \textbf{0.00\%} / 1.42 &  0.00\% / 0.02   \\ 
\bottomrule
\end{tabular}%
}\caption{The results of controllability measured on the MACSum$_{Doc}$ dataset. Surprisingly, GTE achieves near-zero failure rates across all attributes with only a few iterations. The bold denotes the best performance. Failure or reaching the maximum number of iterations is denoted as $\circlearrowright$.}\label{tab: main_doc}
\end{table*}

%% file: fig/tab-main_dial.tex
\begin{table*}[]
\centering
\resizebox{1\textwidth}{!}{%
\begin{tabular}{l|cccc|cccc|c|c}
\toprule
\multirow{2}{*}{Model} & \multicolumn{4}{c|}{Extractiveness ($\downarrow$ / $\downarrow$)} & \multicolumn{4}{c|}{Length ($\downarrow$ / $\downarrow$)} & \multirow{2}{*}{Topic ($\downarrow$ / $\downarrow$)} & \multirow{2}{*}{Speaker ($\downarrow$ / $\downarrow$)} \\ \cmidrule(lr){2-5} \cmidrule(lr){6-9} 
 & normal  & high  & fully  & avg  & short  & normal  & long  & avg  & &   \\ 
\midrule
Llama3-70B-\texttt{Iter} &  31.78\% / 8.13 &  43.59\% / 8.40 &  8.16\% / 5.39 & 29.63\% / 7.59 &  12.00\% / $\circlearrowright$ &  93.75\% / 6.00 &  98.00\% / $\circlearrowright$ &   81.79\% / 6.00 &  0.00\% / 0.01 &   0.00\% / 0.00 \\ 
Llama3-70B-\texttt{SAI} &  14.41\% / 5.91 &  23.08\% / 5.31 &  \textbf{0.00\%} / 3.72 &  13.27\% / 5.50 &  \textbf{0.00\%} / 1.25 &  62.05\% / 5.70 &  92.00\% / 9.33 &  57.10\% / 5.62 &  0.00\% / 0.02 &  0.00\% / 0.00 \\ 
Llama3-70B-GTE &  \textbf{0.00\%} / 2.31 &  \textbf{0.00\%} / 2.56 &  4.08\% / 3.64 &  \textbf{0.61\%} / 2.49 &   \textbf{0.00\%} / 1.00 &  \textbf{36.61\%} / 4.73 &  \textbf{80.00\%} / 5.70 &  \textbf{37.65\%} / 4.53 &  0.00\% / 0.01 &  0.00\% / 0.00  \\ 
\midrule
GPT-4o-\texttt{Iter} &  79.24\% / 4.36 &  82.05\% 3.67 &  59.18\% / 1.00 &  76.54\% / 4.00 &  6.00\% / $\circlearrowright$ &  98.21\% / $\circlearrowright$ & 100.00\% / $\circlearrowright$ &  84.26\% / $\circlearrowright$ &  0.31\% / 0.01 &  0.00\% / 0.00 \\ 
GPT-4o-\texttt{SAI} &  84.75\% / 4.00 &  87.18\% 1.50 &  53.06\% 5.10 &  80.25\% / 4.32 &  2.00\% / 4.50 &  96.43\% / $\circlearrowright$ &  100.00\% / $\circlearrowright$ &  82.41\% / 4.50 &  0.00\% / 0.01 &  0.00\% / 0.00 \\
GPT-4o-GTE  &  \textbf{17.80\%} / 7.94 &   \textbf{25.64\%} / 7.92 &  \textbf{8.16\%} / 4.58 &  \textbf{17.28\%} / 7.53 &  \textbf{0.00\%} / 1.40 &  \textbf{9.82\%} / 2.75 &   \textbf{44.00\%} / 4.21 &  \textbf{13.58\%} / 2.90 &  0.00\% / 0.02 &  0.00\% / 0.00 \\ 
\bottomrule
\end{tabular}%
}\caption{The results of controllability measured on the MACSum$_{Dial}$ dataset.}\label{tab: main_dial}
\end{table*}

%% file: content/experiment.tex
\section{Experimental setup}

We evaluate the controllability of various LLMs, including Phi-3-{\small{medium}} \cite{abdin2024phi}, the Llama3 series \cite{dubey2024llama}, and the GPT series \cite{brown2020language, achiam2023gpt}. To analyze model performance by size, we utilize both the 8B \footnote{meta-llama/Meta-Llama-3-8B-Instruct} and quantized 70B versions\footnote{casperhansen/llama-3-70b-instruct-awq} of Llama3, as well as GPT-3.5 and GPT-4o. We use BERTScore \cite{zhang2020bertscore} to measure embedding similarity.
We used two datasets for our experiments: MACSum$_{Doc}$ and MACSum$_{Dial}$~\cite{zhang-etal-2023-macsum}, which comprise committee meeting transcripts and news content, respectively. Both datasets are designed for mixed-attribute summarization that controls multiple attributes simultaneously. Notably, only MACSum$_{Dial}$ include \textit{speaker} attribute. Since we evaluate LLM performance on individual attributes, we use attributes separately.

%% file: content/result.tex
\section{Results and Discussions}

\paragraph{Main results}

\input{fig/fig-length_change}

We denote the naive iteration approach, which repeatedly adjusts attributes, as \texttt{Iter}.
The strategy that provides step-by-step attribute identification is denoted as \texttt{SAI}\textemdash{}a tool-augmented, stronger version of self-correction that provides the correct attribute values, since LLMs struggle to measure summary attributes on their own.
As shown in Table \ref{tab: main_doc}, our GTE demonstrates remarkably lower failure rates and requires fewer iterations when adjusting summaries across all attributes, including challenging numerical attributes in MACSum$_{Doc}$. Surprisingly, GTE reduced the failure rate to nearly 0\% when applied to Llama3-70B and GPT-4o, successfully controlling both \textit{extractiveness} and \textit{length} within just 1–3 iterations. 
For smaller models such as Phi-3-{\small{medium}} and Llama3-8B, which initially exhibited high failure rates, our approach significantly reduced those rates, demonstrating its effectiveness across different model scales. 
In particular, for long \textit{length}\textemdash{}the most challenging attribute\textemdash{}GTE achieved a remarkably low failure rate of just 2.32\% within an average of 3.14 iterations.

LLMs encounter greater difficulty with the MACSum$_{Dial}$ dataset (Table \ref{tab: main_dial}). The dataset, derived from QMSum \cite{zhong-etal-2021-qmsum}, consists of lengthy and diverse content from parliamentary and committee meetings, making it more challenging than the CNN-news-based MACSum$_{Doc}$. 
Notably, length control proved to be the most challenging attribute in MACSum$_{Dial}$. This challenge is likely due to the dataset's origin in long parliamentary transcripts, which makes it inherently difficult to generate summaries of a specific target length. 
While the model handled short-length summaries relatively well, difficulty increased significantly as the requested summary length grew. In fact, for long-length summaries, both GPT-4o-\texttt{Iter} and GPT-4o-\texttt{SAI} showed a 100\% failure rate. 
However, GTE showed meaningfully improved length controllability. With GPT-4o, the average failure rate dropped below 50\%. Notably, for normal-length summaries, the failure rate further reduced from over 90\% to  9.82\%.
Regarding \textit{extractiveness}, the \texttt{Iter} and \texttt{SAI} of GPT-4o exhibit relatively low iteration counts, as the models often exceed the maximum iteration. While their failure rates were close to 80\%, GTE achieved a markedly lower failure rate at 17.28\% with low iterations, demonstrating the effectiveness of our framework.

\paragraph{Gradual change across iteration steps} To analyze how the attribute changes at each step, we track \textit{length} adjustments per iteration (Figure \ref{fig: length_change}). 
While all methods start with a similar distribution at the initial point, GTE consistently converges within approximately three iterations, maintaining a stable length adjustment pattern across samples. 
In contrast, \texttt{Iter} and \texttt{SAI} show inconsistent changes across samples, resulting in higher variance in length adjustments. 
This demonstrates that our method enables robust attribute control with fewer iterations, regardless of the data sample.
For this experiment, we use Llama3-70B and randomly select 110 samples from the MACSum$_{Doc}$ test set.

\paragraph{Attribute types} We observe that LLMs control linguistic attributes (\textit{topic} and \textit{speaker}) better than numerical attributes (\textit{extractiveness} and \textit{length}). This aligns with previous research in mathematical reasoning, where LLMs struggle with numerical features \cite{akhtar-etal-2023-exploring}, highlighting a broader challenge in precisely handling numerical constraints.
From the perspective of the summarization task, \textit{extractiveness} and \textit{length} control the structure of the summary, whereas \textit{topic} and \textit{speaker} influence its content. Our findings suggest that LLMs are proficient at adjusting content to align with user preferences but struggle to generate summaries with specific structural constraints.

\input{fig/tab-quality}

\paragraph{Quality of controlled summary}

We evaluate the quality of generated summaries including UniEval \cite{zhong-etal-2022-towards} and QuestEval \cite{scialom-etal-2021-questeval}, as they correlate highly with human judgments and assess the overall quality of the summary itself. UniEval is a multi-dimensional evaluator that assesses \textit{coherence}, \textit{consistency}, \textit{fluency}, and \textit{relevance} of summaries. QuestEval measures precision and recall by leveraging a question-answering framework to compare the content between the source document and the generated summary without relying on the reference summary.
Table \ref{tab: quality} shows that our method's summaries outperform across all UniEval dimensions and QuestEval, demonstrating effective attribute control while maintaining overall summary quality. 
While \texttt{Iter} and \texttt{SAI} generate misaligned summaries with lower \textit{relevance} (i.e., how well a summary retains key information compared to the reference), GTE effectively aligns them, resulting in a substantial gain.

Although our primary objective is to control the summary rather than match the reference, we report ROUGE~\cite{lin-2004-rouge} and BERTScore~\cite{zhang2020bertscore} to ensure comprehensive evaluation, as they are representative metrics in text summarization. 
GTE achieves higher scores than other iterative approaches, demonstrating across various evaluation metrics that GTE not only enhances controllability but also improves overall summary quality.

%% file: fig/fig-length_change.tex
\begin{figure*}[h]
\centering
\includegraphics[width=1\linewidth]{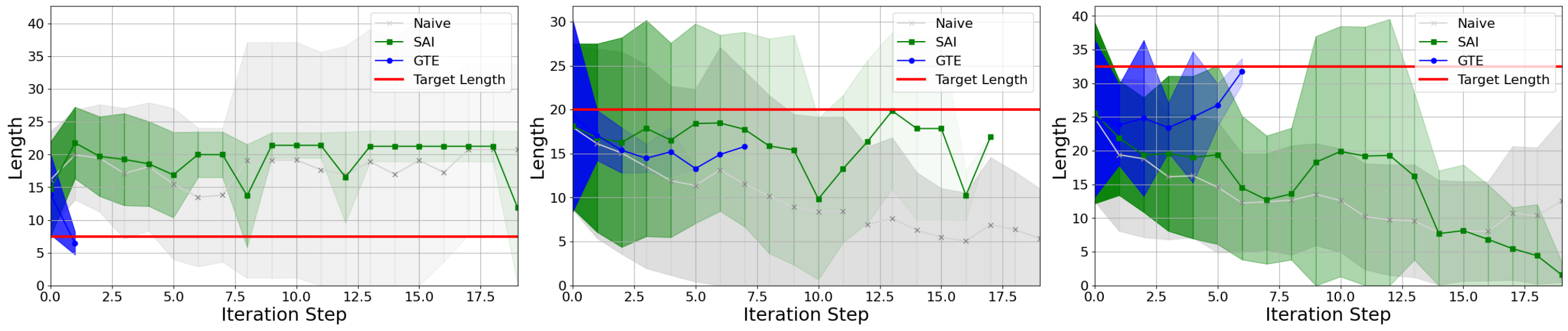}
\caption{The graphs show how the length ratio changes for each iteration. The intensity of the distribution color is proportional to the number of data points, and the markers represent the average values. The red line indicates the target length, with values of 7.5\%, 20\%, and 32.5\% from left to right.}
\label{fig: length_change}
\end{figure*}


%% file: fig/tab-quality.tex
\begin{table*}
\centering
\scalebox{0.8}{
\begin{tabular}{l|ccccc|c|c|c}
\toprule
\multirow{2}{*}{Model} & \multicolumn{5}{c|}{UniEval} & \multirow{2}{*}{QuestEval} & \multirow{2}{*}{BERTScore} & \multirow{2}{*}{ROUGE-1} \\ 
 \cmidrule(lr){2-6} 
& Coherence & Consistency  & Fluency  & Relevance & Overall & & \\
\midrule
\texttt{Iter} (Ext) & 0.820 & 0.800 & 0.859 & 0.696 & 0.794 & 0.523 & 0.826 & 0.194\\
\texttt{SAI} (Ext) & 0.884 & 0.843 & 0.905 & 0.785 & 0.864 & 0.554 & 0.848 & 0.229 \\
\texttt{Iter} (Len) & 0.836 & 0.803 & 0.836  & 0.759 & 0.808 & 0.484 & 0.829 & 0.235 \\
\texttt{SAI} (Len) & 0.934 & 0.834 & 0.942  & 0.887 & 0.899 & 0.548 & 0.867 & 0.270 \\
\midrule
GTE (Ext) & \textbf{0.941} & \textbf{0.873} & 0.937 & 0.880 & \textbf{0.908} & \textbf{0.590} & 0.861 & 0.236 \\
GTE (Len) & 0.937 & 0.840 & \textbf{0.944}  & \textbf{0.901} & 0.905 & 0.553 & \textbf{0.868} & \textbf{0.272} \\
\bottomrule
\end{tabular}
}\caption{Among the iterative methods, GTE demonstrates both effective attribute control and noticeable improvements in summary quality.}\label{tab: quality}
\end{table*}

%% file: content/discussion.tex
\section{Mixed attributes}
We extend our evaluation to assess whether LLMs can precisely handle mixed-attribute control.
While models manage to control linguistic attributes, they struggle with numerical attributes. 
Simultaneous control over all attributes remains challenging for all iterative methods, including GTE.
In GTE, the model is guided to identify the causes of its errors and regenerate summaries by incorporating this feedback. However, in a mixed attribute setting, the model must process multiple instances of \texttt{SAI} and \texttt{SEG} for each attribute simultaneously, increasing the cognitive load and making precise control of all attributes more difficult.
Therefore, unlike single-attribute evaluation\textemdash{}which assesses whether individual attributes are accurately controlled\textemdash{}we evaluate mixed-attribute control by measuring errors using mean absolute deviation (MAD). This approach compares the differences between the attributes of the generated summary and the requested values, providing a more flexible evaluation of attribute control.

\paragraph{Sequential-planning}

Recognizing the challenges in precisely controlling all attributes simultaneously, we introduce a sequential planning strategy, \textit{min-planning}, which gradually adjusts attributes\textemdash{}starting with those that are most poorly controlled in the initial draft\textemdash{}using GTE.
Figure \ref{fig: mixed} shows the results comparing single-attribute control with iterations to mixed-attribute control using \textit{min-planning} on the MACSum$_{Doc}$ dataset. We refer to the initial summary in the mixed-attribute control setting as the \textit{mixed-draft}.
The \textit{min-planning} method shows a modest improvement in controlling both attributes compared to the \textit{mixed-draft}. However, attributes are still not fully controlled as in single-attribute models, highlighting the difficulty of balancing multiple attributes.
We anticipate that modifying one attribute often disrupts previously adjusted attributes due to underlying correlations.
For example, even if \textit{length} is adjusted first, it may still change when \textit{extractiveness} is subsequently controlled.
Additionally, \textit{min-planning} adjusts each attribute only once without iteration, which may explain its inability to fully control the attributes. A single refinement is often insufficient, whereas GTE iteratively regenerates the summary until the target attribute is successfully adjusted in single-attribute control.
Exploring ways for LLMs to control multiple attributes simultaneously would be promising future work.

\input{fig/fig-mixed}

%% file: fig/fig-mixed.tex
\begin{figure}
\centering
\includegraphics[width=0.92\linewidth]{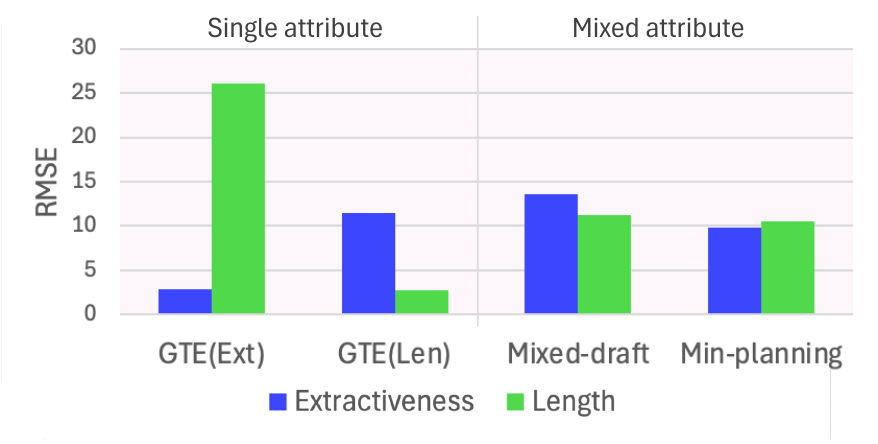}
\caption{Correlations among attributes hinder LLMs’ ability to control them jointly in mixed-attribute setting.}
\label{fig: mixed}
\end{figure}

%% file: content/conclusion.tex
\section{Conclusion}
In this work, we systematically explore the controllability of LLMs.
To this end, we revisit the measurement of summary attributes.  
We evaluate the controllability of LLMs via iterative assessment and find that they struggle more with numerical attributes than linguistic ones. To address this, we propose GTE, a self-correction-based approach that first instructs the model to identify misaligned attributes and then guides it to explain its own error to generate better-controlled summaries in subsequent iterations. GTE enables LLMs to control challenging numerical attributes with lower failure rates and fewer iterations. Furthermore, we demonstrate the high quality of controlled summaries using various evaluation metrics.

%% file: content/limitation.tex
\section*{Limitation}

We explore the controllability of various attributes in LLMs and introduced a novel guide-to-explain (GTE) framework to address challenges in numerical attributes. While GTE enhanced successfully control over challenging numerical attributes, it still struggled with highly correlated mixed numerical attributes. Additionally, \textit{min-planning}, which adjusts attributes in order of least alignment, also faced difficulties achieving precise control. Even after properly adjusting one attribute, modifying the correlated numerical attribute caused the previously adjusted attribute to change. We believe further research could explore more effective methods for addressing these challenges.

%% file: content/ethics.tex
\section*{Ethics}

We used publicly available MACSum datasets for our research, conducting experiments with Phi-3, Llama3 \footnote{Meta Llama3 Community License, Copyright © Meta Platforms, Inc. All Rights Reserved. More details can be found at: \href{https://llama.meta.com/llama3/license/}{Llama3 License}}, GPT-3.5, and GPT-4o from April to October 2024. 

%% file: content/acknowledge.tex
\section*{Acknowledgments}

This work was supported by the National Research Foundation of Korea (NRF) grant funded by the Korea government (MSIT) (No. RS-2023-00217286) (45\%); by Culture, Sports and Tourism R\&D Program through the Korea Creative Content Agency grant funded by the Ministry of Culture, Sports and Tourism in 2025 (No: RS-2025-02413038, Development of an AI-Based Korean Diagnostic System for Efficient Korean Speaking Learning by Foreigners) (45\%); and by Institute of Information \& communications Technology Planning \& Evaluation (IITP) grant funded by the Korea government (MSIT) (No.RS-2019-II191906, Artificial Intelligence Graduate School Program (POSTECH)) (10\%)

%% file: content/appendix.tex


\input{fig/tab-attribute}

\appendix

\section{Attribute details}\label{appendix:attribute_distribution}

Table~\ref{tab: data_distribution} presents the distributions of the MACSum$_{Doc}$ and MACSum$_{Dial}$ training datasets used in our study. For each attribute, we report the distribution of attribute values corresponding to each assigned label, with the average shown in parentheses. 
For \textit{extractiveness}, both datasets show a wide range of values within each label but exhibit similar average values: around 85\% for \textsf{normal}, 90\% for \textsf{high}, and 100\% for \textsf{fully}. These averages are used as the relabeled target values.
For \textit{length}, the observed averages deviate from the annotation guide. In MACSum$_{Doc}$, the means are 4.6\% (\textsf{short}), 6.9\% (\textsf{normal}), and 13.9\% (\textsf{long}), while in MACSum$_{Dial}$, they are 2.0\%, 3.7\%, and 6.0\%, respectively. Due to the small gaps between label means, relabeling based on these values would not sufficiently capture LLM controllability for length. Therefore, we follow the annotation guide and relabel with target values of 7.5\% (\textsf{short}), 15\% (\textsf{normal}), and 32.5\% (\textsf{long}).
For \textit{topic}, both datasets show similar scores. We consider summaries with scores above the minimum threshold of 74 to be topic-focused. Similarly, for \textit{speaker}, we use a minimum threshold of 75, derived from the distribution of reference summaries, to define speaker-focused outputs.



\section{Guide-to-explain (GTE) prompts}\label{appendix:gte_prompts}

Below, we present the prompts and example outputs used for each attribute within the GTE framework.

\clearpage

\input{fig/fig-appendix_length}

\input{fig/fig-appendix_ext}

\input{fig/fig-appendix_topic}

\label{sec:appendix}

%% file: fig/tab-attribute.tex
\begin{table*}[t]
    \centering
    \resizebox{0.90\textwidth}{!}{%
    \begin{tabular}{llcccccc}
        \toprule
        \multirow{2}{*}{\textbf{Attribute}} & \multirow{2}{*}{\textbf{Label}} & \multicolumn{3}{c}{\textbf{MACSum$_{Doc}$}} & \multicolumn{3}{c}{\textbf{MACSum$_{Dial}$}} \\
        \cmidrule(lr){3-5} \cmidrule(lr){6-8}
         &  & \textbf{Distributions} & \textbf{Relabel} & \textbf{\# of summaries} & \textbf{Distributions} & \textbf{Relabel} & \textbf{\# of summaries} \\
         \midrule
        \multirow{3}{*}{\textit{Extractiveness}} 
        & \textsf{normal} & 35.7 - 100.0\% (85.2\%) & 85.0\% & 3731 & 53.2 - 100.0\% (86.4\%)  & 85.0\% & 1661 \\
        & \textsf{high} & 55.0 - 100.0\% (90.0\%)  & 90.0\% & 287 & 63.0 - 100.0\% (88.9\%) & 90.0\% & 340 \\
        & \textsf{fully} & 84.6 - 100.0\% (99.7\%)  & 100.0\% & 260 & 75.9 - 100.0\% (98.4\%) & 100.0\% & 337 \\
        \midrule
        \multirow{3}{*}{\textit{Length}} 
        & \textsf{short} & 0.7 - 15.0\% (4.8\%) & 7.5\% & 1059 & 0.2 - 20.8\% (2.0\%) & 7.5\% & 300 \\
        & \textsf{normal} & 0.5 - 48.6\% (6.9\%) & 20.0\% & 2194 & 0.3 - 41.9\% (3.7\%) & 20.0\% & 1693 \\
        & \textsf{long} & 1.5 - 39.8\% (13.9\%) & 32.5\% & 1025 & 0.7 - 32.4\% (6.0\%) & 32.5\% & 345 \\
        \midrule
        \textit{Topic} & - & 74.8 - 88.8 & 74.0 & 2013 & 73.6 - 87.0 & 74.0 & 2317 \\
        \midrule
        \textit{Speaker} & - & - & - & - & 75.6 - 92.0 & 75.0 & 1796 \\
        \bottomrule
    \end{tabular}}
    \caption{Data distributions of MACSum$_{Doc}$ and MACSum$_{Dial}$.}\label{tab: data_distribution}
\end{table*}

%% file: fig/fig-appendix_length.tex
\begin{figure*}
\centering
\includegraphics[width=0.95\linewidth]{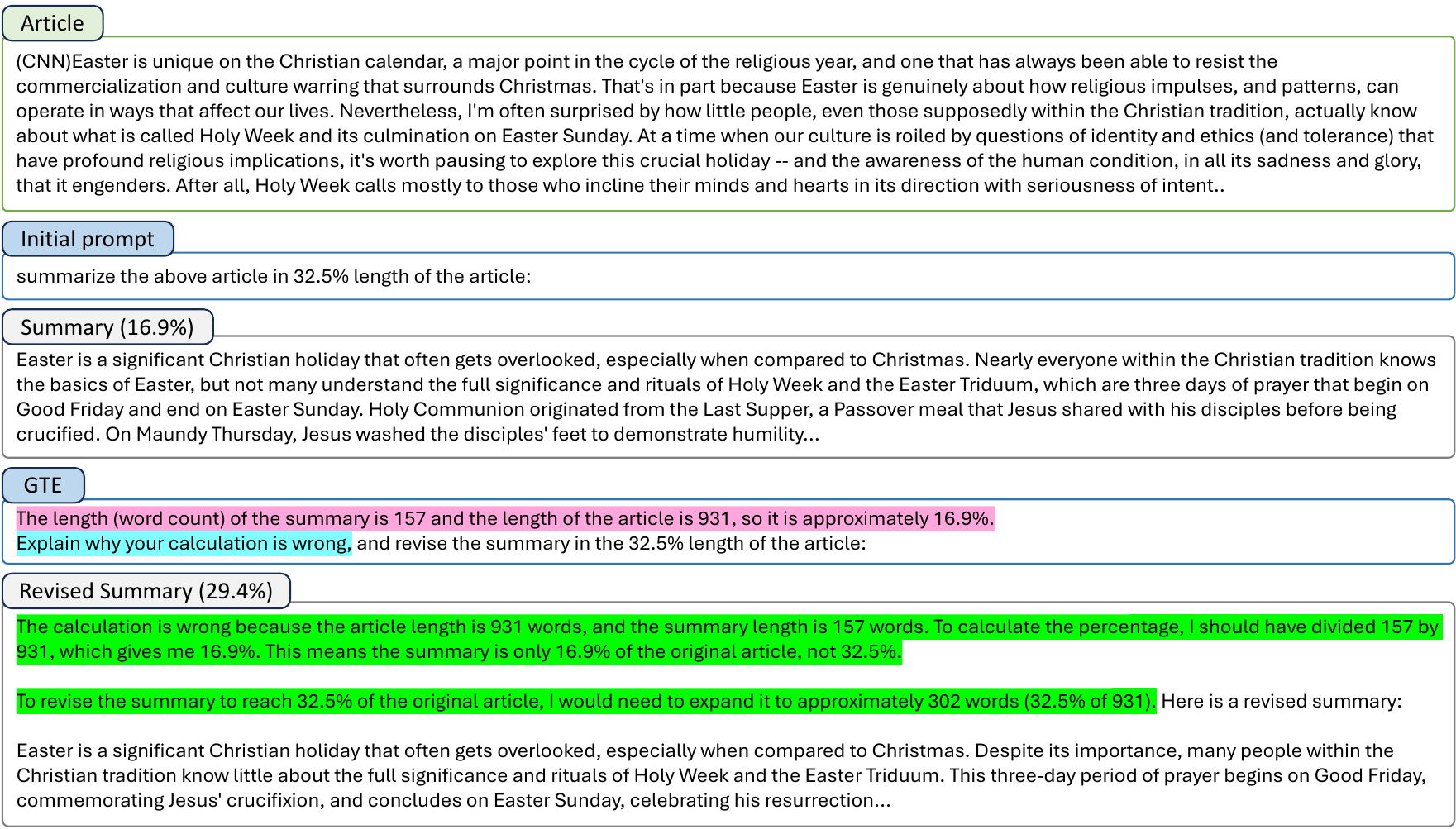}
\caption{Length guide-to-explain (GTE).}
\label{fig: appendix-length1}
\end{figure*}

\begin{figure*}
\centering
\includegraphics[width=0.95\linewidth]{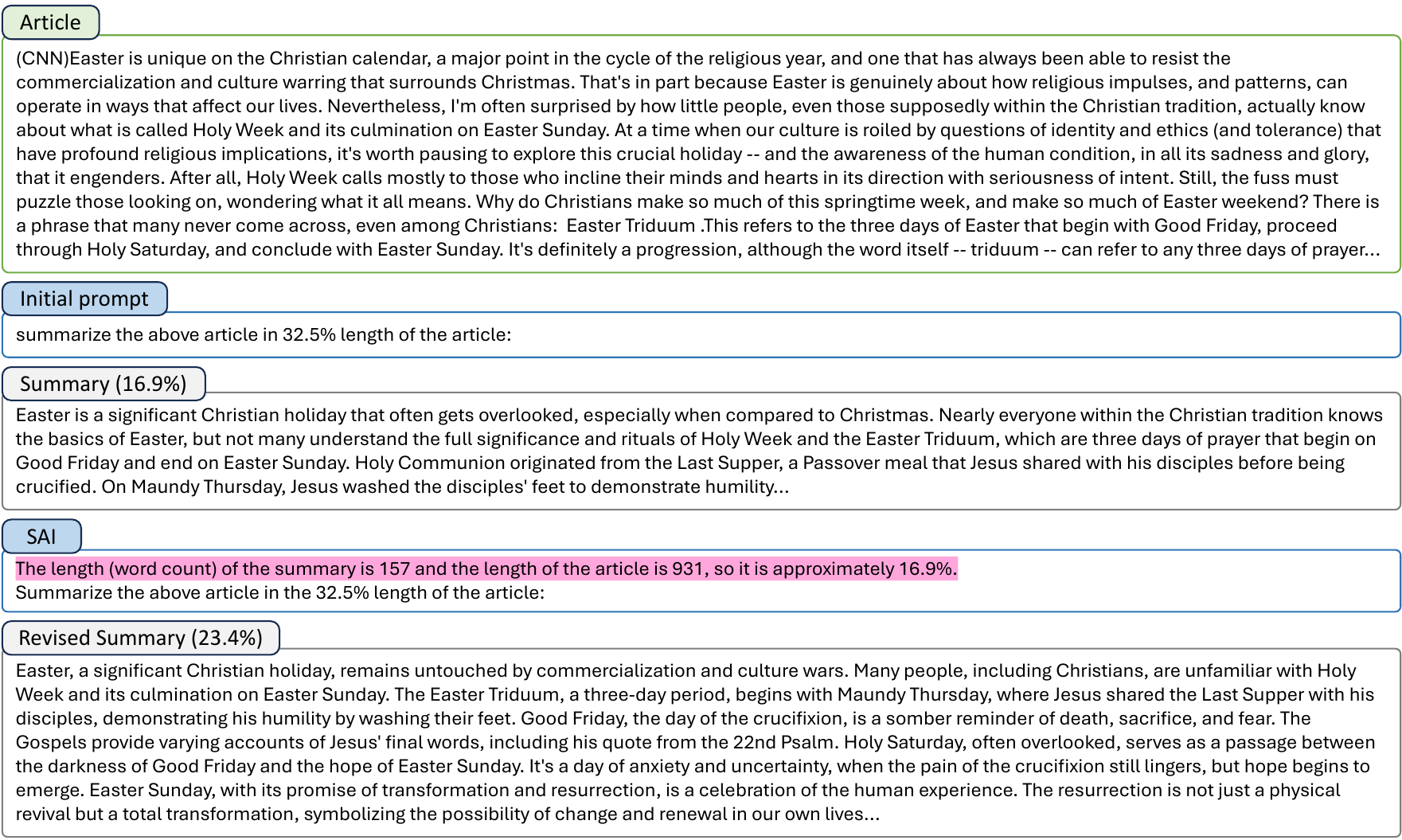}
\caption{Length attribute-identification (\texttt{SAI}).}
\label{fig: appendix-length2}
\end{figure*}

\begin{figure*}
\centering
\includegraphics[width=0.95\linewidth]{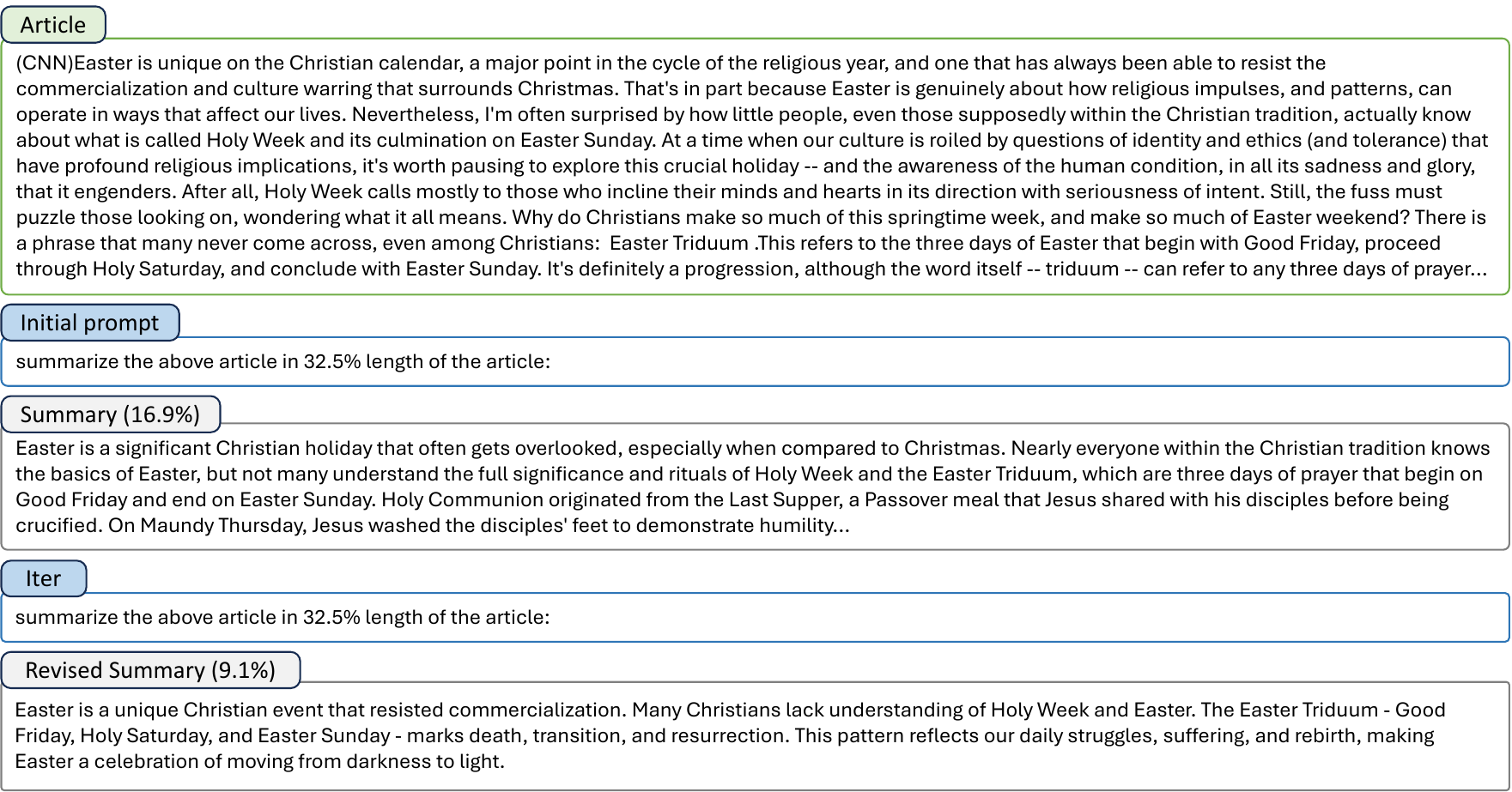}
\caption{Length iteration (\texttt{Iter}).}
\label{fig: appendix-length3}
\end{figure*}

%% file: fig/fig-appendix_ext.tex
\begin{figure*}
\centering
\includegraphics[width=0.95\linewidth]{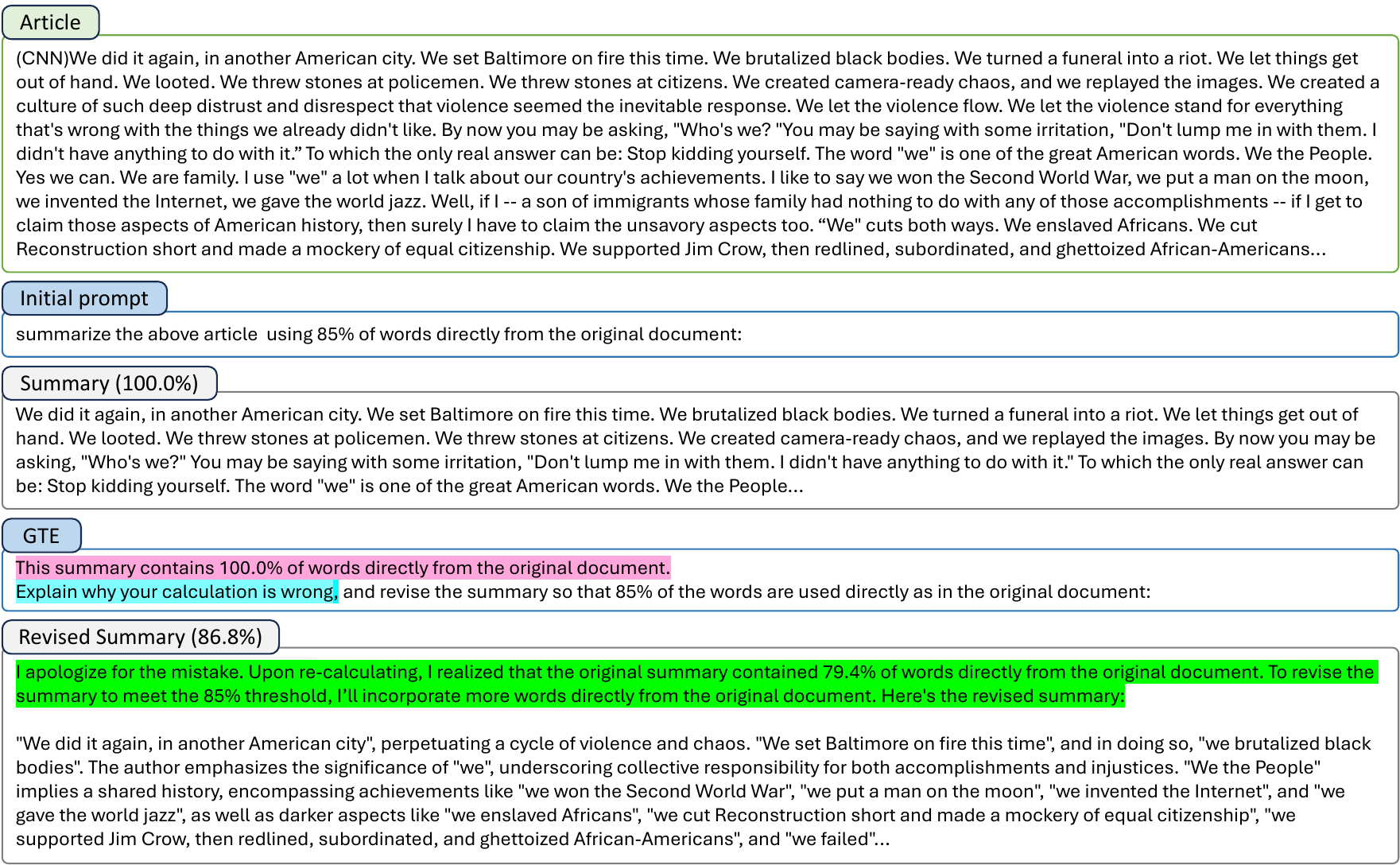}
\caption{Extractiveness guide-to-explain (GTE).}
\label{fig: appendix-ext1}
\end{figure*}

\begin{figure*}
\centering
\includegraphics[width=0.95\linewidth]{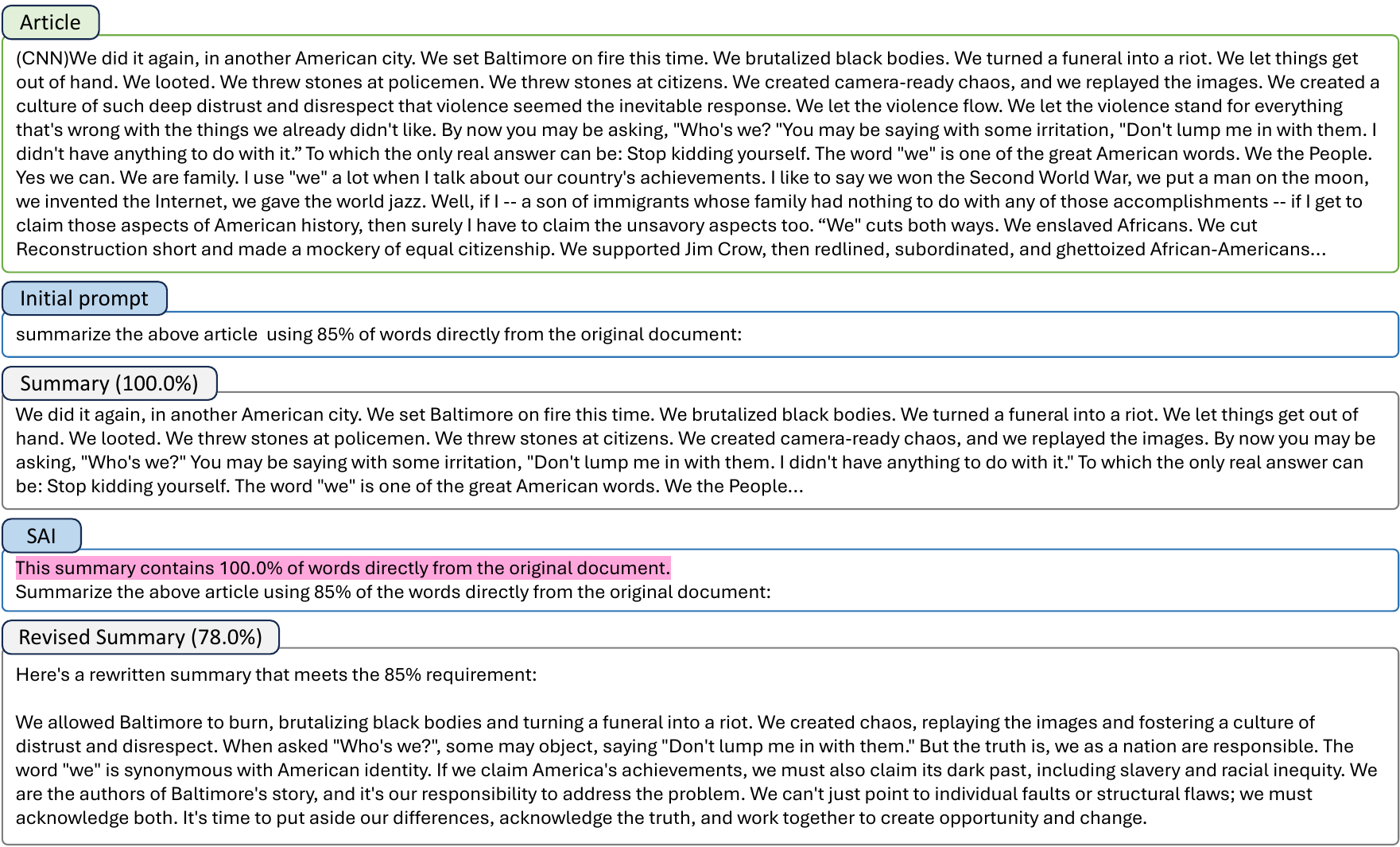}
\caption{Extractiveness attribute-identification (\texttt{SAI}).}
\label{fig: appendix-ext2}
\end{figure*}

\begin{figure*}
\centering
\includegraphics[width=0.95\linewidth]{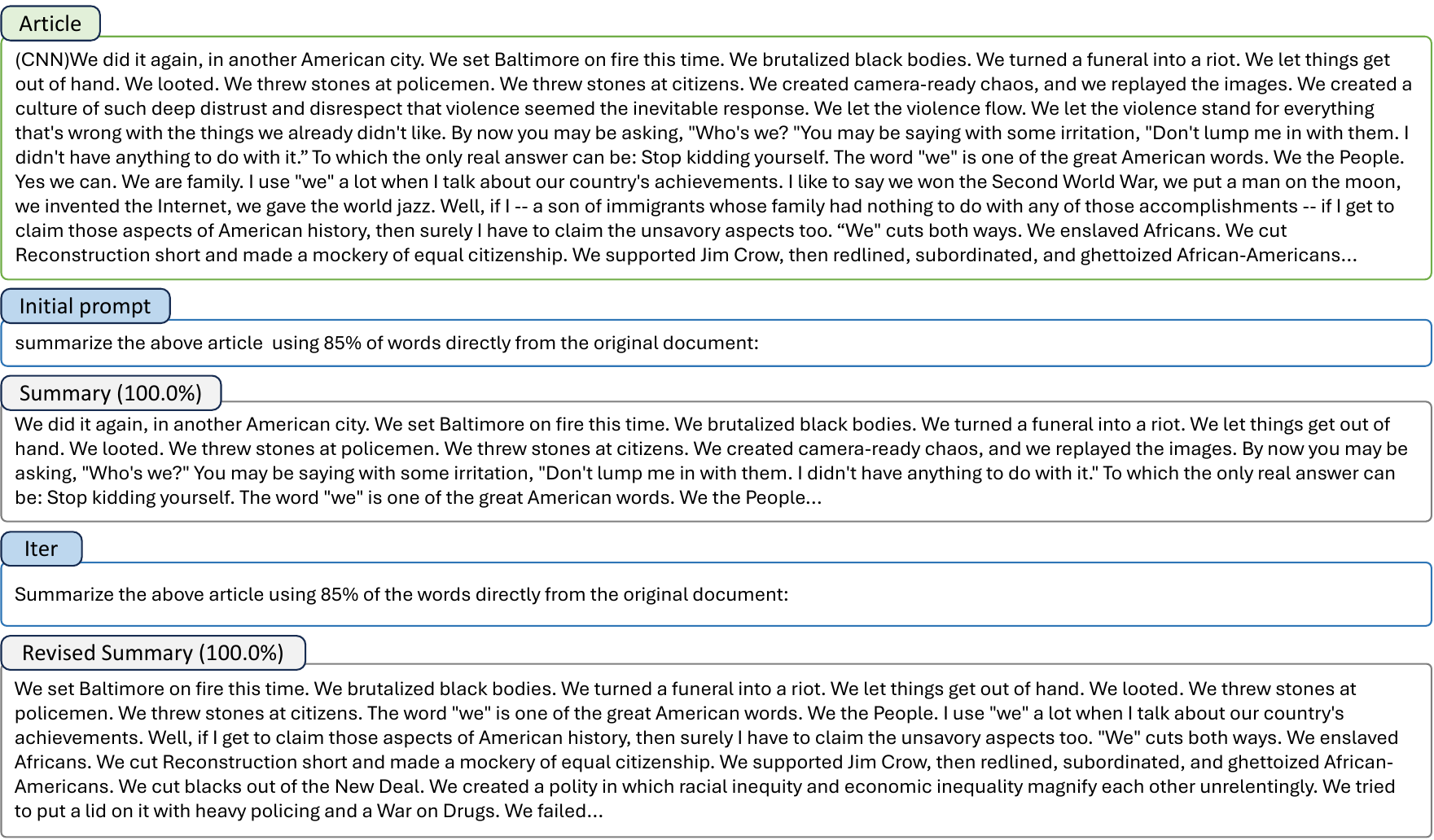}
\caption{Extractiveness iteration (\texttt{Iter)}.}
\label{fig: appendix-ext3}
\end{figure*}

%% file: fig/fig-appendix_topic.tex
\begin{figure*}
\centering
\includegraphics[width=0.95\linewidth]{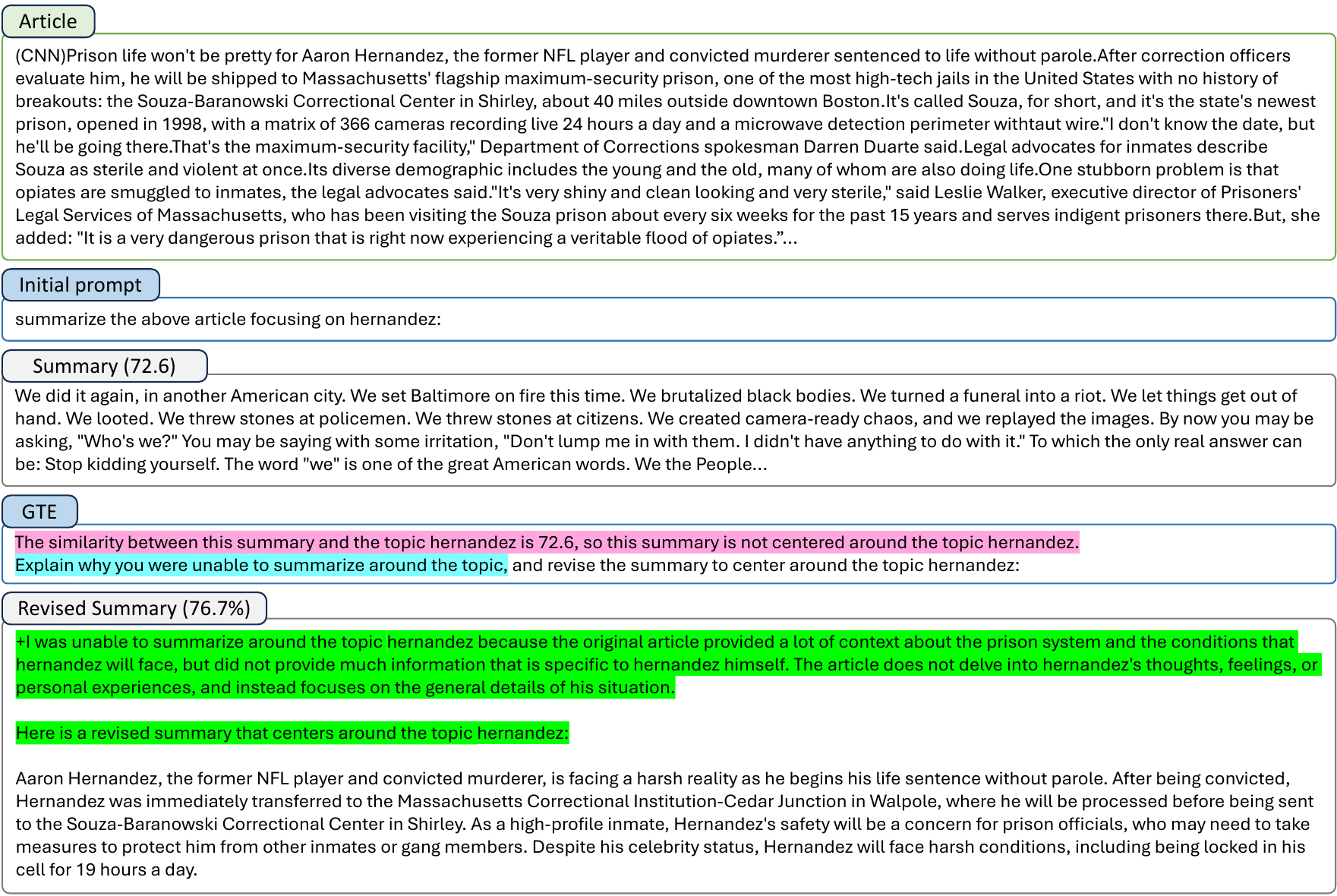}
\caption{Topic guide-to-explain (GTE).}
\label{fig: appendix-topic1}
\end{figure*}